\documentclass[conference]{IEEEtran}
\usepackage{times}

\usepackage[numbers]{natbib}
\usepackage{multicol}

\usepackage{epsfig}
\usepackage{graphicx}
\usepackage{amssymb}

\usepackage{booktabs}
\usepackage{color}
\usepackage{amsmath,bm}
\usepackage{multirow}
\usepackage{float}
\usepackage{subcaption} 
\usepackage{tabularray}
\usepackage{array}
\usepackage{pifont}
\let\labelindent\relax
\usepackage{enumitem}
\usepackage{colortbl}
\usepackage{algorithm}
\usepackage{makecell}
\usepackage{float} 
\usepackage{comment}
\usepackage{adjustbox}

\usepackage[bookmarks=true,pagebackref,breaklinks,colorlinks]{hyperref}

\begin{document}

\title{\textbf{\emph{WalnutData}}: A UAV Remote Sensing Dataset of Green Walnuts and Model Evaluation}

\author{Mingjie Wu$^{1,2*}$, Chenggui Yang$^{1,2*}$, Huihua Wang$^{1,2}$, Chen Xue$^{1,2}$, Yibo Wang$^{1,2}$, Haoyu Wang$^{1,2}$ \\
Yansong Wang$^{1,2}$, Can Peng$^{1,2}$, Yuqi Han$^{1,2}$, Ruoyu Li$^{1,2}$, Lijun Yun$^{1,2,3\dagger}$, Zaiqing Chen$^{1,2}$, Yuelong Xia$^{1}$ \\

$^1$School of Information, Yunnan Normal University, \\
$^2$Engineering Research Center of Computer Vision and Intelligent Control Technology, \\
Department of Education of Yunnan Province, \\
$^3$Southwest United Graduate School  \\
$^*$Equal Contribution, $^{\dagger}$Corresponding Author \\
}

\twocolumn[
{%
\renewcommand\twocolumn[1][]{#1}
\maketitle
}]
\begin{abstract}
The UAV technology is gradually maturing and can provide extremely powerful support for smart agriculture and precise monitoring. Currently, there is no dataset related to green walnuts in the field of agricultural computer vision. Thus, in order to promote the algorithm design in the field of agricultural computer vision, we used UAV to collect remote-sensing data from 8 walnut sample plots. Considering that green walnuts are subject to various lighting conditions and occlusion, we constructed a large-scale dataset with a higher-granularity of target features - WalnutData. This dataset contains a total of 30,240 images and 706,208 instances, and there are 4 target categories: being illuminated by frontal light and unoccluded (A1), being backlit and unoccluded (A2), being illuminated by frontal light and occluded (B1), and being backlit and occluded (B2). Subsequently, we evaluated many mainstream algorithms on WalnutData and used these evaluation results as the baseline standard. The dataset and all evaluation results can be obtained at \url{https://github.com/1wuming/WalnutData}. 
\end{abstract}

\begin{figure}
	\centering
	\includegraphics[width=1\linewidth]{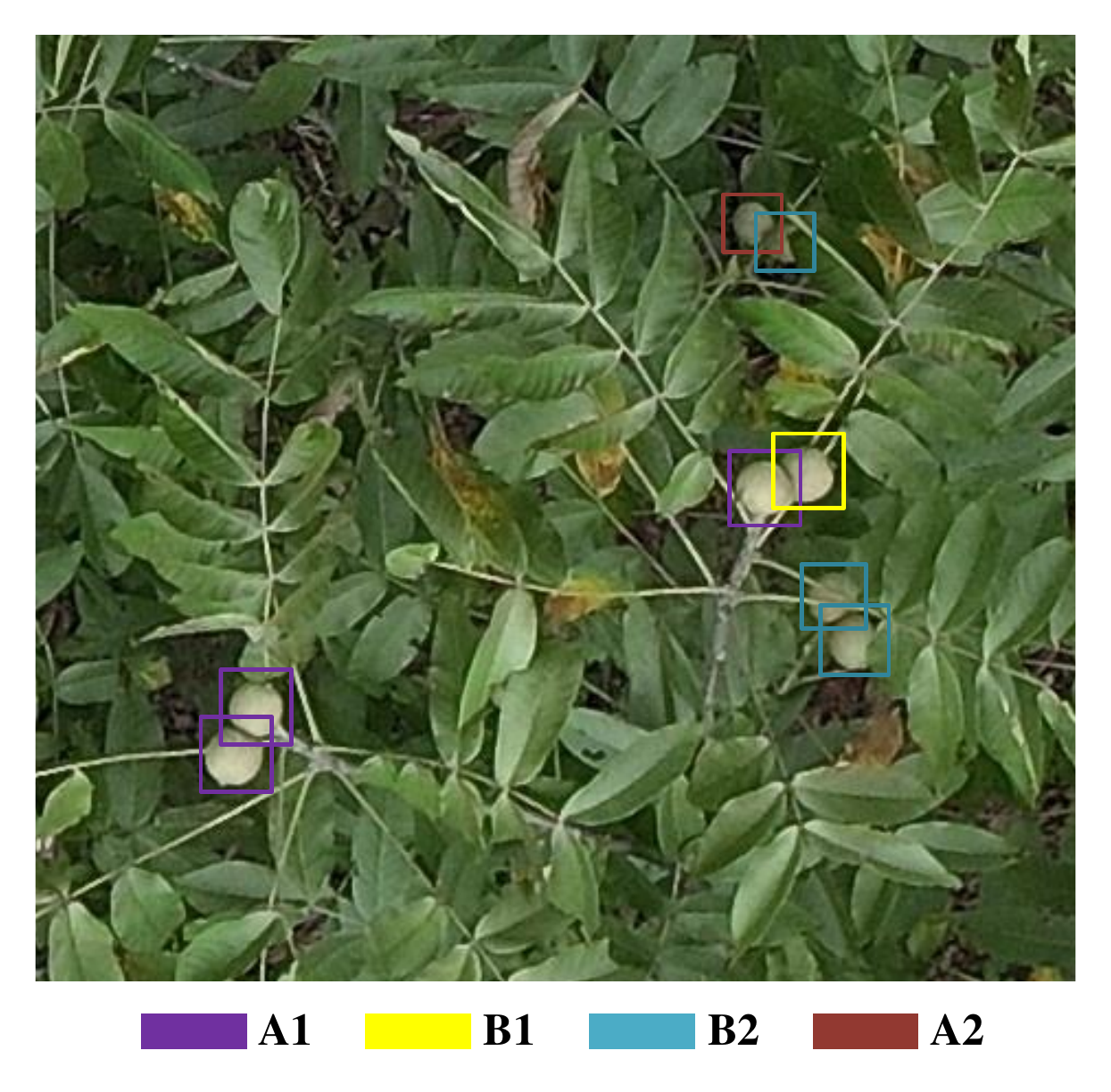}
	\caption{Examples of local image categories in the WalnutData. Category A1 represents green walnuts that are illuminated by frontal light and unoccluded. Category B1 represents green walnuts that are illuminated by frontal light and occluded. Category B2 represents green walnuts that are backlit and obstructed. Category A2 represents green walnuts that are backlit and unobstructed.}
	\label{fig:Examples of WalnutData Categories}
\end{figure}

\IEEEpeerreviewmaketitle
\section{Introduction} ~\label{Sec:Introduction}
Currently, UAV technology is approaching maturity and is sufficient to provide reliable assistance in fields such as agroforestry production management~\cite{jia2024maize,zheng2024robust}, emergency rescue~\cite{quero2025unmanned,wen2024route}, and security monitoring~\cite{zhu2025multiscale}. In the production of agricultural and forestry crops, UAV, by carrying multi-modal or high-resolution camera sensors, can quickly acquire image information of large-scale farmland and orchards, providing strong technical support for crop detection, yield estimation, and automated management in precision agriculture~\cite{ariza2024object}. Therefore, UAV technology is widely applied in the agricultural field. However, object detection combined with UAV technology faces many challenges, such as lighting changes, foliage occlusion, and the diversity of target scales~\cite{liang2025enhanced,du2023dsw,wu2024walnut}. These problems significantly increase the detection difficulty and limit the performance of existing algorithms in practical applications.

As a crop of great value~\cite{yang2025walnut}, the green walnut has a complex surface texture, a high color similarity to the background, and is often affected by the occlusion of branches, leaves, and lighting changes. These characteristics make it a research object with unique scientific challenges and engineering application value in agricultural object detection. In the future of the smart walnut industry, in automated applications such as aerial UAV picking robots, accurate object detection is not only the basis for crop positioning but also the core prerequisite for robot path planning, obstacle avoidance decision-making, and picking priority judgment. If the impact of environmental interference on the apparent characteristics of the target is ignored, the reliability of the detection algorithm will directly affect the efficiency and success rate of the robot's task execution. Therefore, data-driven automated management methods for walnut production will greatly need a large-scale dataset with higher-granularity target features.

Currently, most UAV-based object detection datasets are related to urban road environments, such as VisDrone~\cite{zhu2021detection} and UAVDT~\cite{du2018unmanned}, or datasets for maritime object detection, such as SeaDronesSee~\cite{varga2022seadronessee} and SDS-ODv2~\cite{kiefer20231st}. There are only relatively few open-source datasets for UAV-based object detection of agricultural and forestry crops. In addition, most of the object detection datasets related to agricultural crops are obtained by shooting with mobile phones or hand-held cameras, such as MinneApple~\cite{hani2020minneapple} and TomatoPlantfactoryDataset~\cite{wu2023dataset}. Therefore, this study aims to construct a large-scale walnut dataset obtained from UAV aerial photography and make it open-source worldwide.

Although UAV technology provides an efficient means of data collection for the object detection of agricultural and forestry crops, existing research mostly regards agricultural and forestry crops as a single category, ignoring the differences in apparent characteristics caused by environmental interference, such as backlight, frontal light, and occlusion. Although this simplification can improve the detection accuracy in the short term, it is difficult to meet the fine-grained perception requirements of picking robots for target states, thus severely restricting the development of their autonomous capabilities. Solving the detection problem of walnut fruits can provide a transferable technical paradigm for the automated management of other agricultural crops such as citrus and apples.

Therefore, to solve the above-mentioned series of current problems and meet the requirements of the smart walnut field, this study introduces the first large-scale UAV low-altitude remote-sensing green walnut object detection dataset - \textbf{WalnutData}. This dataset includes a total of 30,240 RGB images with a resolution of 1,024×1,024 pixels, and the total number of annotated instances is as high as 706,208. It innovatively divides the targets into four environmental states, including A1 (being illuminated by frontal light and unoccluded), A2 (being backlit and unoccluded), B1 (being illuminated by frontal light and occluded), and B2 (being backlit and occluded), as shown in Fig.~\ref{fig:Examples of WalnutData Categories}. The main contributions of this research are as follows:
\begin{itemize}[left=0pt]
	\item  As far as we know, the WalnutData is the largest green walnut object detection dataset with annotation labels in the field of agricultural computer vision. This dataset refines the lighting and occlusion conditions of walnut fruits and has multiple categories. It can be used for the further development of object detectors in automated applications within the intelligent walnut production management.
	
	\item We used the WalnutData to conduct benchmark tests on the current mainstream one-stage detection algorithms such as DETR and the YOLO series, as well as two-stage detection algorithms like Fast R-CNN and Faster R-CNN. These algorithms can serve as the basis for future algorithm design.
\end{itemize}

\section{Related Works} ~\label{Sec:related_works}
\begin{table*}
	\caption{Comparison between WalnutData and other aerial datasets.}
	\label{tab:Table_1}
	\setlength{\tabcolsep}{12pt}
	\centering
	\begin{tabular}{lccccccc}
		\toprule
		Dataset & Environment & Image Widths & Altitude Range & Angle Range & Images & Instances & Classes \\
		\midrule
		VisDrone~\cite{zhu2021detection} & Traffic & 960-2,000 & 5-200m & 0-90° & 8,599 & 540,000 & 10 \\
		UAVDT~\cite{du2018unmanned} & Traffic & 1,024 & 5-200m & 0-90° & 80,000 & 840,000 & 3 \\
		SeaDronesSee~\cite{varga2022seadronessee} & Maritime & 3,840-5,456 & 5-260m & 0-90° & 8,295 & - & 6 \\
		SDS-ODv2~\cite{kiefer20231st} & Maritime & 3,840-5,456 & 5-260m & 0-90° & 14,227 & 403,192 & 6 \\
		\textbf{WalnutData(our)} & Agriculture & 1,024 & 12-30m & 90° & 30,240 & 706,208 & 4 \\
		\bottomrule
	\end{tabular}
\end{table*}

\begin{table*}
	\caption{Compare WalnutData with the annotated datasets in agriculture in recent years.}
	\label{tab:Table_2}
	\centering
	\begin{tabular}{lcccccc}
		\toprule
		Dataset & Object & Device & Images & Classes & Image Widths & Label Type \\
		\midrule
		Apeinans I et al.~\cite{apeinans2024cherry} & Cherry & - & 2283 & 1 & 640 & YOLO \\
		Wu Z et al.~\cite{wu2023dataset} & Tomato & DSLR camera and mobile phone & 520 & 2 & 4,000/4,032 & VOC \\
		Hani N et al.~\cite{hani2020minneapple} & Apple & Mobile phone & 1,001 & 1 & 1280 & - \\
		Bargoti S et al.~\cite{bargoti2017deep} & Apple,mango, and almond & UGV and DSLR camera & 2,750 & 3 & 202/300/500 & COCO \\
		Stein M et al.~\cite{stein2016image} & Mango & UGV & 1,500 & 1 & 3,296 & - \\
		Santos T et al.~\cite{santos2021methodology} & Apple & UAV & 1,139 & 1 & 256 & COCO \\
		Butte S et al.~\cite{butte2021potato} & Potato & UAV & 360 & 2 & 1,500 & VOC \\
		\textbf{WalnutData(our)} & Walnut & UAV & 30,240 & 4 & 1,024 & VOC, COCO, YOLO \\
		\bottomrule
	\end{tabular}
\end{table*}
In this section, we review the main annotated datasets that can be used for supervised learning models in the field of UAV vision and agricultural scenarios.
\subsection{Annotated Image Datasets Collected by UAV}
In recent years, most of the mainstream annotated datasets collected by UAV are used to describe data of traffic roads or marine environments, such as VisDrone ~\cite{zhu2021detection}, UAVDT~\cite{du2018unmanned}, SeaDronesSee~\cite{varga2022seadronessee}, and SDS-ODv2~\cite{kiefer20231st}. As can be seen from Table~\ref{tab:Table_1}, the images captured by these UAVs have a height range of 5-260 meters, a shooting angle range of 0-90°, and an image width range of 960-5,456 pixel. These datasets cover various scenarios such as cities, villages, and oceans, mainly focusing on road traffic environment analysis and maritime rescue, but lacking coverage of agricultural target scenarios. WalnutData is an agricultural scenario dataset different from the traffic and maritime fields. The UAV images are collected in the range of 12-30m, with an aerial shooting angle of -90°. The width of the dataset images is 1,024 pixel after the original images are segmented. WalnutData has a significant advantage over other datasets in terms of the number of images and instances.

\subsection{Annotated Datasets in Agriculture}
With the rapid popularization of deep learning technology and the urgent needs of precision agriculture, more and more datasets in the field of computer vision for agriculture have been constructed and made public. The main objects of study in these datasets listed in Table~\ref{tab:Table_2} include apple, potato, tomato, mango, etc. However, there is still a lack of research on green walnut targets.

In studies such as tomato~\cite{wu2023dataset}, apple~\cite{bargoti2017deep}, and mango~\cite{stein2016image}, the image data are mainly collected by DSLR camera, UGV, or mobile phone. These shooting methods are affected by the ground environment. Moreover, the planting terrains of crops such as tomato, cherry, or apple are relatively flat, which is quite different from the growing terrain of walnut trees. In addition, in the research on agricultural UAV-related datasets, the apple trees studied by Santos T et al.~\cite{santos2021methodology} have a neat interval, which is very conducive to collecting relatively regular data information. Thus, effective algorithms can be used for apple detection, tracking, and positioning. Butte S et al.~\cite{butte2021potato} proposed a potato dataset. Through the model they designed, it is possible to accurately identify healthy or drought-stressed potatoes, providing a new idea for precision agriculture.

In Yunnan Province, China, most walnut trees are planted in mountainous areas with large altitude differences and complex terrains, and the fruit trees are unevenly distributed~\cite{wang2025ow}. Therefore, in this study, UAV aerial photography is used for data collection to obtain WalnutData. Compared with other datasets, WalnutData has a more detailed division of crop characteristic states and a larger amount of data, which can provide a more solid foundation for model design. In addition, WalnutData provides three types of labels (VOC, COCO, and YOLO), which are suitable for many current mainstream object detection models and offer multiple choices for researchers in related fields.

\section{WalnutData Construction}
\begin{table*}
	\caption{The detailed data of the walnut sample plots in this study. A total of 8 sample plots were selected, all of which are located in Yangbi County, Dali Bai Autonomous Prefecture, Yunnan Province, China. The shooting dates are between July 18 and September 14, 2024. In order to capture the changes in lighting conditions, the shooting time was chosen between 9:00 and 19:00. At the same time, in order to minimize the impact on the quality of the images collected by the sensor as much as possible, we set the flight altitude between 12 and 30 meters.}
	\label{tab:Table_3}
	\centering
	\begin{tabular}{lcccccccc}
		\toprule
		\multirow{2}{*}{Sample Number} & \multirow{2}{*}{Altitude Range (m)} & \multicolumn{2}{c}{Geographical Coordinates} & \multirow{2}{*}{Flight Altitude (m)} & \multirow{2}{*}{GSD (cm/pixel)} & \multirow{2}{*}{Date} & \multirow{2}{*}{Time} & \multirow{2}{*}{Images} \\
		\cmidrule(lr){3-4}
		& & E & N & & & & & \\
		\midrule
		1 & 2031.48-2033.77 & 100°0'25.600" & 25°40'9.077" & 25 & 0.31 & 2024/7/18 & 9:49 & 251 \\
		2 & 2062.44-2068.22 & 100°1'37.681" & 25°40'47.933" & 12 & 0.15 & 2024/7/18 & 11:20 & 3703 \\
		3 & 1871.56-1882.45 & 100°1'29.779" & 25°36'37.303" & 15 & 0.19 & 2024/8/31 & 10:54 & 828 \\
		4 & 1905.41-1905.47 & 100°1'14.148" & 25°36'54.632" & 20 & 0.25 & 2024/8/30 & 18:25 & 656 \\
		5 & 2339.12-2339.73 & 99°52'6.610" & 25°36'28.995" & 25 & 0.31 & 2024/7/20 & 16:03 & 691 \\
		6 & 2092.40-2096.17 & 99°48'29.425" & 25°38'15.829" & 20 & 0.25 & 2024/9/1 & 10:00 & 1503 \\
		7 & 2131.30-2131.38 & 100°1'53.728" & 25°40'18.837" & 30 & 0.38 & 2024/9/13 & 10:05 & 236 \\
		8 & 2045.32-2064.72 & 100°1'43.099" & 25°40'8.263" & 30 & 0.38 & 2024/9/14 & 11:04 & 1531 \\
		\bottomrule
	\end{tabular}
\end{table*}

\subsection{Data Collection}
We carried out data collection on 8 walnut sample plots between July 18 and September 14, 2024. These sample plots are all located in Yangbi County, Dali Bai Autonomous Prefecture, Yunnan Province, China. In addition, in order to capture the changes in lighting conditions, we conducted the shooting between 9:00 and 19:00. The data collection equipment used uniformly was a DJI Matrice 300 RTK UAV and a Zenmuse P1 (35mm F2.8) lens. The UAV took photos from a top-down angle (-90°) along the pre-planned flight path throughout the process, and the flight path fully covered the scope of each sample plot. To reduce the impact of too high a flight altitude and too fast camera movement on the imaging quality, and while ensuring flight safety, we set the flight speed between 1-3 m/s and the flight altitude between 12-30 m. The information of the walnut sample plots selected in this study is shown in Table~\ref{tab:Table_3}.

Finally, we set the overlap rate of the UAV flight paths to be all above 70\%. A total of 9,399 images were collected from the 8 walnut sample plots. 
\subsection{Dataset Construction}
Setting the overlap rate of the flight paths above 70\% can ensure that certain contents will not be missed during shooting. However, this will cause the UAV to capture similar areas during the data collection task, resulting in the situation where the same walnut tree appears in multiple aerial images. In order to avoid a large amount of duplicate content in the final dataset, we organized multiple members to carefully screen the aerial images of each walnut sample plot at the same time, so as to achieve the situation where there are almost no overlapping areas in the selected images.

Since the resolution of the UAV aerial images (8,192×5,460 pixels) is too large, which is not conducive to the training of the model, in this study, the selected original images were all cut with a step size of 512. The resolution of the cut images is 1,024×1,024 pixels. After the processing of the above steps, the dataset of this study was finally formed, with a total of 30,240 images.

\subsection{Data Annotation}
In this study, four label categories were defined: A1 (frontal light without occlusion), A2 (backlight without occlusion), B1 (frontal light with occlusion), and B2 (backlight with occlusion). The Labelme annotation tool was used to manually annotate the dataset, and the annotation format is bounding box. During this work process, we organized multiple members to spend about 3 months on data annotation, and finally obtained 24,673 labels.

\subsection{Dataset Split}
\begin{figure}[t]
	\centering
	\includegraphics[width=0.9\linewidth]{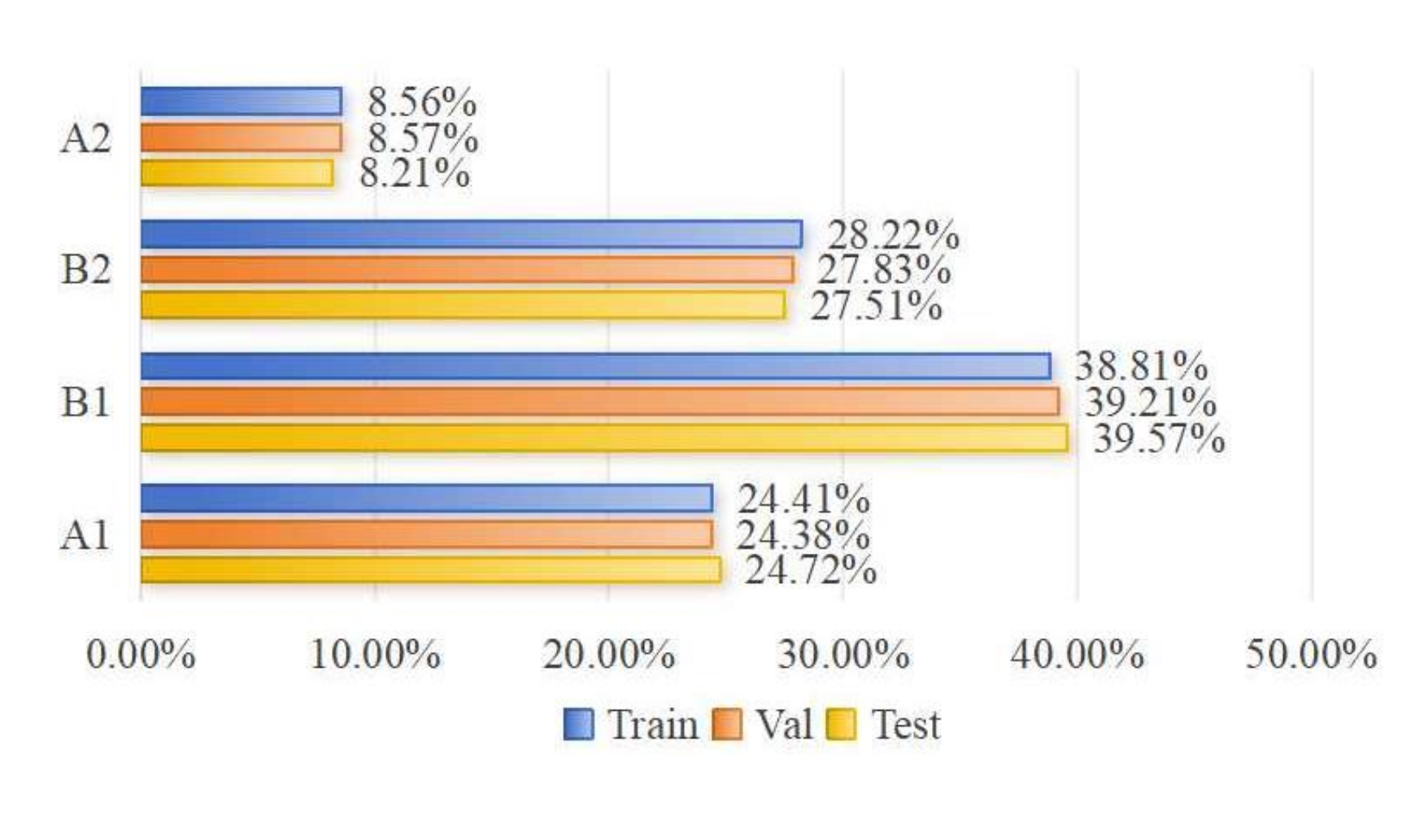}
	\caption{The proportion information of the number of instances in each category after the dataset is partitioned. The proportions of the numbers of A1, B1, B2, and A2 instances are similar in the Train, the Val, and the Test respectively.}
	\label{fig:Instance Distribution Information after Dataset Division}
\end{figure}

According to the way accepted by the current mainstream object detection models, we divided the dataset into a Train, a Val, and a Test. The ratio of the Train, the Val, and the Test is 7:2:1, with 21,167 images, 6,048 images, and 3,025 images respectively. In addition, in the arrangement of the distribution of the number of categories, we tried our best to ensure the similarity and balance of the distribution. The distribution information of category instances after the dataset partition is shown in Fig.~\ref{fig:Instance Distribution Information after Dataset Division}. We will release the Train and the Val containing label annotations. At the same time, the Test will also be provided to researchers for evaluating their own models, but the label annotations of the Test images will not be provided.

\subsection{Dataset Analysis}
We have counted the number of instances and the average number of instances in WalnutData (Table~\ref{tab:Table_4}). The average number of targets per image in the Training, the Val, and the Test is approximately 23.353.

\setlength{\tabcolsep}{8pt} 
\begin{table}[htbp]
	\centering
	\caption{The distribution of the number of instances in WalnutData and the average number of bounding boxes per image. BBox is the abbreviation of Bounding Box, and Avg. BBox quantity represents the average number of bounding boxes per image.}
	\label{tab:Table_4}
	\begin{tabular}{lccc}
		\toprule
		Name & Image quantity & BBox quantity & Avg. BBox quantity \\
		\midrule
		Train & 21,167 & 495,812 & 23.424 \\
		Val & 6,048 & 139,255 & 23.025 \\
		Test & 3,025 & 71,141 & 23.518 \\
		\bottomrule
	\end{tabular}
\end{table}

\begin{figure}
	\centering
	\includegraphics[width=1.05\linewidth]{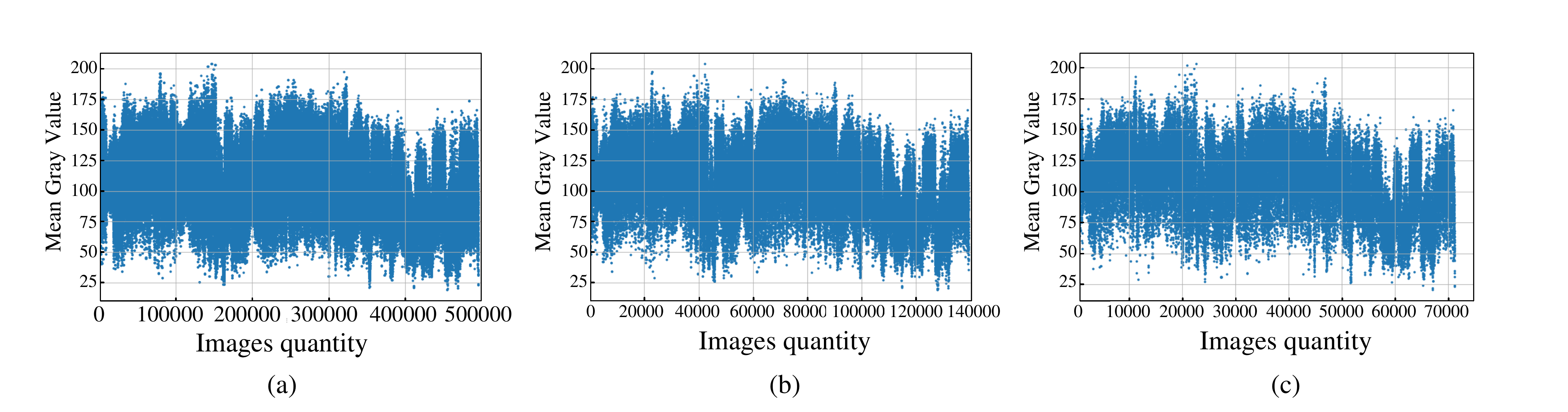}
	
	\caption{The distribution of the average grayscale values of each instance in the Train, the Val, and the Test. (a), (b), and (c) are the statistics of the grayscale values of each instance in the Train, the Val, and the Test respectively. Among them, 76.31\%, 75.59\%, and 75.81\% of the instances in the Train, the Val, and the Test respectively have grayscale values lower than the median grayscale value (127.5), indicating that more than half of the green walnuts in WalnutData receive less light. }
	\label{fig:Grayscale value information}
\end{figure}

We analyzed the lighting conditions of the green walnut fruits in WalnutData. Since the lighting conditions of the non-target backgrounds around the green walnut fruits are almost similar, we first extracted the pixels of the images within the instance rectangular boxes. Then, we converted the RGB images into grayscale images and calculated the average grayscale value to analyze the lighting intensity received by the green walnut fruits. The distribution of the average grayscale values of each instance in the Train, the Val, and the Test is shown in Fig.~\ref{fig:Grayscale value information}. The average grayscale values of the Train, the Val, and the Test are 107.316, 108.048, and 107.544 respectively. The proportions of values lower than the intermediate grayscale value of 127.5 are 76.31\%, 75.59\%, and 75.81\% respectively. This indicates that most of the green walnuts in WalnutData are in backlight conditions or are shaded by leaves in relatively dark places. 

In addition, according to the definition of large (pixel\textgreater96), medium (96\textgreater pixel\textgreater32), and small (32\textgreater pixel) targets in the COCO dataset~\cite{lin2014microsoft}, we counted the quantity distribution of large, medium, and small targets in WalnutData (Table~\ref{tab:Table_5}). In WalnutData, the proportion of medium and small targets is higher, which is in line with the morphological characteristics of green walnut fruits from the perspective of a UAV. Therefore, the model trained on WalnutData can better adapt to the distribution of target sizes in practical application scenarios. 

\setlength{\tabcolsep}{15pt} 
\begin{table}[htbp]
	\centering
	\caption{Quantity distribution of large, medium and small targets in WalnutData.}
	\label{tab:Table_5}
	\begin{tabular}{lccc}
		\toprule
		Name & Large & Medium & Small \\
		\midrule
		Train & 526 & 211,669 & 283,617 \\
		Val & 175 & 59,371 & 79,709 \\
		Test & 74 & 30,047 & 41,020 \\
		\bottomrule
	\end{tabular}
\end{table}

\setlength{\tabcolsep}{10pt} 
\begin{table}
	\centering
	\caption{The Val benchmark evaluation results of one-stage object detection algorithms in the Ultralytics framework for WalnutData. The top two results in the evaluation are represented by bold and underline respectively. The top two in the mAP50 metric are YOLOv3 (95.1\%) and YOLOv3-SPP (94.9\%) respectively, and the top two in the mAP50:95 metric are YOLOv8x (72.7\%) and YOLOv10x (72.6\%) respectively.}
	\label{tab:Table_6}
	\begin{tabular}{lcccc}
		\toprule
		Method & Size & GFLOPs & mAP50  & mAP50:95  \\
		\midrule
		\multirow{3}{*}{YOLOv3~\cite{redmon2018yolov3}} 
		& - & 154.6 & \underline{94.9} & 71.6 \\
		& SPP & 155.4 & \textbf{95.1} & 71.4 \\
		& Tiny & 12.9 & 66.2 & 38.0 \\
		\midrule
		YOLOv4~\cite{bochkovskiy2020yolov4} & - & - & 57.3 & 31.3 \\
		\midrule
		\multirow{5}{*}{YOLOv5~\cite{jocher2022ultralytics}} 
		& n & 4.1 & 72.0 & 45.1 \\
		& s & 15.8 & 84.5 & 57.0 \\
		& m & 47.9 & 90.9 & 64.7 \\
		& l & 107.7 & 93.3 & 68.4 \\
		& x & 203.8 & 94.5 & 70.8 \\
		\midrule
		\multirow{4}{*}{YOLOv6~\cite{li2022yolov6}} 
		& n & - & 74.2 & 47.8 \\
		& s & - & 87.8 & 59.6 \\
		& m & - & 83.7 & 56.9 \\
		& l & - & 87.1 & 60.1 \\
		\midrule
		YOLOv7~\cite{wang2023yolov7} & - & - & 67.0 & 40.1 \\
		\midrule
		\multirow{5}{*}{YOLOv8~\cite{jocher2022ultralytics}} 
		& n & 8.1 & 75.2 & 49.2 \\
		& s & 28.4 & 86.2 & 59.7 \\
		& m & 78.7 & 92.2 & 68.0 \\
		& l & 164.8 & 93.6 & 70.6 \\
		& x & 257.4 & 94.6 & \textbf{72.7} \\
		\midrule
		\multirow{5}{*}{YOLOv9~\cite{wang2024yolov9}} 
		& t & - & 72.2 & 47.3 \\
		& s & - & 80.9 & 55.3 \\
		& m & - & 89.7 & 64.1 \\
		& c & - & 92.1 & 67.8 \\
		& e & - & 93.8 & 70.6 \\
		\midrule
		\multirow{6}{*}{YOLOv10~\cite{wang2025yolov10}} 
		& n & 8.2 & 75.9 & 49.6 \\
		& s & 24.5 & 86.3 & 59.9 \\
		& m & 63.4 & 89.1 & 63.4 \\
		& b & 98.0 & 90.9 & 65.9 \\
		& l & 126.3 & 92.0 & 67.6 \\
		& x & 169.8 & 94.4 & \underline{72.6} \\
		\midrule
		\multirow{5}{*}{YOLOv11~\cite{khanam2024yolov11}} 
		& n & 6.3 & 74.6 & 48.5 \\
		& s & 21.3 & 84.7 & 58.7 \\
		& m & 67.7 & 91.3 & 66.7 \\
		& l & 86.6 & 91.7 & 68.0 \\
		& x & 194.4 & 94.0 & 71.7 \\
		\bottomrule
	\end{tabular}
\end{table}

\setlength{\tabcolsep}{4.5pt} 
\begin{table}
	\centering
	\caption{The Val benchmark evaluation results of one-stage object detection algorithms in the mmdetection framework for WalnutData. The top two results in the evaluation are represented by bold and underline respectively. The top two in the AP50:95 metric are YOLOX-x (54.9\%) and YOLOX-l (51.6\%) respectively, the top two in the AP50 metric are YOLOX-x (82.7\%) and YOLOX-l (79.1\%) respectively, and the top two in the AP75 metric are YOLOX-x (67.5\%) and YOLOX-l (61.7\%) respectively.}
	\label{tab:Table_7}
	\begin{tabular}{lccccccc}
		\toprule
		Method           &  Size  &  Backbone  &  AP  & AP50 & AP75 &  \\ \midrule
		\multirow{3}{*}{YOLOX~\cite{ge2021yolox}} &   s    & CSPDarknet & 45.4 & 72.8 & 52.9 &  \\
		&   l    & CSPDarknet & \underline{51.6} & \underline{79.1} & \underline{62.7} &  \\
		&   x    & CSPDarknet & \textbf{54.9} & \textbf{82.7} & \textbf{67.5} &  \\ \midrule
		DETR~\cite{carion2020end}          &   -    &  ResNet50  & 14.1 & 34.7 & 8.0  &  \\
		Deformable DETR~\cite{zhu2020deformable}     &   -    &  ResNet50  & 49.2 & 76.8 & 59.7 &  \\
		DINO~\cite{zhang2022dino}         & 4scale &  ResNet50  & 50.3 & 77.0 & 61.7 &  \\
		Conditional DETR~\cite{meng2021conditional}    &   -    &  ResNet50  & 37.9 & 65.5 & 40.9 &  \\ \bottomrule
	\end{tabular}
\end{table}

\setlength{\tabcolsep}{12pt} 

\begin{table*}
	\centering
	\caption{The Val benchmark evaluation results of the one-stage object detection algorithm under the Ultralytics framework for each category in the WalnutData.  The top two results in the evaluation are represented by bold and underline respectively. In terms of the mAP50 metric, YOLOv3-SPP demonstrates the best detection performance under conditions such as occlusion and backlighting. For the mAP50:95 metric, YOLOv8x shows higher accuracy in the B1 and B2 categories, reaching 73.1\% and 68.8\%, respectively. In the case of no occlusion, regardless of front lighting or backlighting, the difference between YOLOv8x and the first place (YOLOv10x) is only 0.1\%.}
	\label{tab:Table_8}
	\begin{tabular}{lccccccccc}
		\toprule
		
		\multirow{2}{*}{Method} & \multirow{2}{*}{Size} & \multicolumn{4}{c}{mAP50} & \multicolumn{4}{c}{mAP50:95} \\
		
	\cmidrule(lr){3-6} \cmidrule(lr){6-10}
		& & A1 & B1 & B2 & A2 & A1 & B1 & B2 & A2 \\
		\midrule
		\multirow{3}{*}{YOLOv3~\cite{redmon2018yolov3}}
		& - & 96.4 & \underline{96.0} & \underline{94.6} & 92.7 & 75.3 & 71.9 & \underline{68.6} & 70.7 \\
		& SPP & 96.4 & \textbf{96.1} & \textbf{94.7} & \textbf{93.0} & 74.9 & 71.5 & 68.3 & 70.9 \\
		& Tiny & 73.9 & 71.7 & 64.9 & 54.1 & 46.9 & 39.2 & 33.0 & 33.0 \\
		\midrule
		YOLOv4~\cite{bochkovskiy2020yolov4} & - & 74.0 & 54.7 & 49.8 & 40.6 & 44.2 & 32.7 & 23.7 & 24.6 \\
		\midrule
		\multirow{5}{*}{YOLOv5~\cite{jocher2022ultralytics}}
		& n & 80.5 & 78.1 & 68.9 & 60.4 & 54.5 & 47.2 & 39.1 & 39.6 \\
		& s & 89.0 & 87.9 & 84.3 & 76.9 & 63.9 & 58.2 & 52.7 & 53.4 \\
		& m & 93.1 & 92.8 & 90.9 & 86.6 & 69.6 & 65.2 & 61.2 & 62.8 \\
		& l & 94.9 & 94.7 & 93.4 & 90.1 & 72.5 & 68.8 & 65.5 & 66.9 \\
		& x & 95.9 & 95.8 & 94.3 & 91.9 & 74.6 & 71.2 & 68.1 & 69.5 \\
		\midrule
		YOLOv7~\cite{wang2023yolov7} & - & 80.1 & 74.0 & 61.6 & 52.3 & 51.1 & 41.6 & 33.9 & 33.8 \\
		\midrule
		\multirow{5}{*}{YOLOv8~\cite{jocher2022ultralytics}}
		& n & 85.2 & 80.9 & 69.6 & 65.2 & 59.5 & 51.1 & 41.9 & 44.2 \\
		& s & 92.2 & 89.4 & 82.1 & 81.1 & 67.3 & 60.8 & 53.3 & 57.3 \\
		& m & 95.7 & 93.9 & 89.7 & 89.3 & 73.7 & 68.6 & 62.8 & 66.8 \\
		& l & 96.2 & 95.0 & 91.8 & 91.5 & 75.4 & 71.1 & 66.3 & 69.6 \\
		& x & \textbf{96.8} & 95.7 & 93.2 & \underline{92.9} & \underline{77.0} & \textbf{73.1} & \textbf{68.8} & \underline{71.8} \\
		\midrule
		\multirow{5}{*}{YOLOv9~\cite{wang2024yolov9}} 
		& t & 82.7 & 78.5 & 66.6 & 61.0 & 57.7 & 50.0 & 40.1 & 41.4 \\
		& s & 88.4 & 85.5 & 76.8 & 72.9 & 64.1 & 57.3 & 48.7 & 51.0 \\
		& m & 94.3 & 92.0 & 86.5 & 85.8 & 70.9 & 64.8 & 58.3 & 62.3 \\
		& c & 95.8 & 93.8 & 89.6 & 89.2 & 73.7 & 68.4 & 62.6 & 66.4 \\
		& e & \textbf{96.8} & 95.1 & 91.3 & 92.0 & 75.8 & 71.1 & 65.7 & 70.0 \\
		\midrule
		\multirow{6}{*}{YOLOv10~\cite{wang2025yolov10}} 
		& n & 85.2 & 80.8 & 70.6 & 67.0 & 59.3 & 51.1 & 42.4 & 45.4 \\
		& s & 92.1 & 89.2 & 82.5 & 81.6 & 67.5 & 60.6 & 53.6 & 58.0 \\
		& m & 94.0 & 91.6 & 85.8 & 85.1 & 70.1 & 64.2 & 57.6 & 61.7 \\
		& b & 94.8 & 93.0 & 88.1 & 87.8 & 71.7 & 66.5 & 60.7 & 64.6 \\
		& l & 95.5 & 93.6 & 89.3 & 89.4 & 73.0 & 67.9 & 62.6 & 66.7 \\
		& x & \underline{96.7} & 95.5 & 92.6 & 92.6 & \textbf{77.1} & \underline{72.8} & 68.5 & \textbf{71.9} \\
		\midrule
		\multirow{5}{*}{YOLOv11~\cite{khanam2024yolov11}} 
		& n & 84.3 & 79.9 & 69.1 & 65.0 & 58.5 & 50.4 & 41.3 & 43.7 \\
		& s & 91.3 & 88.0 & 80.5 & 79.2 & 66.7 & 59.8 & 52.0 & 56.0 \\
		& m & 95.0 & 93.1 & 88.7 & 88.5 & 72.5 & 67.3 & 61.4 & 65.8 \\
		& l & 95.2 & 93.3 & 89.4 & 88.9 & 73.5 & 68.5 & 63.2 & 66.8 \\
		& x & 96.5 & 95.2 & 92.2 & 91.9 & 76.3 & 72.2 & 67.5 & 70.8 \\
		\bottomrule
	\end{tabular}
\end{table*}

\begin{table*}
	\centering
	\caption{The benchmark evaluation results for small, medium, and large object detection on the WalnutData Val using a one-stage object detection algorithm under the mmdetection framework.  The top two results in the evaluation are represented by bold and underline respectively. YOLOX-x demonstrates the best performance for small and medium-sized object detection, with YOLOX-l ranking second for small objects and Deformable DETR ranking second for medium-sized objects. Additionally, Deformable DETR is more sensitive in detecting large objects, with an accuracy that exceeds the second-place method (DINO) by 7.8\%. The corresponding AR metric also places it in second place (72.5\%).}
	\label{tab:Table_9}
	\setlength{\tabcolsep}{8pt}
	\begin{tabular}{lccccccc}
		\toprule
		Method & Size & AP-small & AP-medium & AP-large & AR-small & AR-medium & AR-large \\
		\midrule
		\multirow{3}{*}{YOLOX~\cite{ge2021yolox}} & s & 44.1 & 46.8 & 37.3 & 64.1 & 64.3 & 45.9 \\
		& l & \underline{50.1} & 53.3 & 44.5 & 67.0 & 67.8 & 55.6 \\
		& x & \textbf{53.5} & \textbf{56.3} & 52.7 & \textbf{68.4} & 68.4 & 58.4 \\
		\midrule
		DETR~\cite{carion2020end} & - & 10.3 & 18.6 & 32.4 & 24.5 & 39.6 & 46.7 \\
		Deformable DETR~\cite{zhu2020deformable} & - & 43.7 & \underline{55.4} & \textbf{64.7} & 59.4 & \underline{69.5} & \underline{72.5} \\
		DINO~\cite{zhang2022dino} & 4scale & 47.5 & 53.5 & \underline{56.9} & \underline{68.1} & \textbf{72.9} & \textbf{72.8} \\
		Conditional DETR~\cite{meng2021conditional} & - & 32.3 & 44.5 & 56.1 & 49.5 & 63.7 & 68.0 \\
		\bottomrule
	\end{tabular}
\end{table*}

\setlength{\tabcolsep}{9pt} 
\begin{table*}
	\centering
	\caption{The benchmark evaluation results for the two-stage object detection algorithms on WalnutData Val. The top two results in the evaluation are represented by bold and underline respectively. In the evaluation results, Cascade R-CNN (ResNet101) demonstrates impressive performance, ranking first in all metrics, with the exception of large object detection, where it ranks second (74.5\%).}
	\label{tab:Table_10}
	\begin{tabular}{lcccccccc}
		\toprule
		Method & Backbone &   AP  & AP50  & AP75 & AP-small  & AP-medium  & AP-large  \\ 
		\midrule
		Fast R-CNN~\cite{girshick2015fast} & ResNet50 &   22.9 & 35.9 & 26.7 & 15.7 & 31.4 & 54.5 \\ 
		Faster R-CNN~\cite{ren2015faster} & ResNet50 &   51.5 & 79.7 & 62.2 & 45.6 & 58.2 & 69.7 \\ 
		Cascade R-CNN~\cite{cai2018cascade} & ResNet50 &  56.0 & 83.7 & 68.4 & 50.3 & 62.4 & 72.6 \\ 
		Grid R-CNN~\cite{lu2019grid} & ResNet50 &   53.8 & 80.4 & 66.1 & 48.2 & 60.2 & 71.9 \\ 
		TridentNet~\cite{li2019scale} & ResNet50 &   53.4 & 80.8 & 64.2 & 48.9 & 58.6 & 69.9 \\ 
		Double head R-CNN~\cite{wu2020rethinking} & ResNet50 &  55.2 & \underline{84.5} & 67.2 & 50.1 & 61.1 & 69.6 \\ 
		Sparse R-CNN~\cite{sun2021sparse} & ResNet50 &   45.3 & 68.8 & 55.8 & 40.6 & 50.9 & 53.8 \\ 
		\midrule
		Fast R-CNN~\cite{girshick2015fast} & ResNet101 &   24.5 & 37.7 & 28.9 & 16.8 & 33.7 & 55.7 \\ 
		Faster R-CNN~\cite{ren2015faster} & ResNet101 & 56.3 & 84.1 & 68.9 & 49.8 & 63.7 & 72.1 \\ 
		Cascade R-CNN~\cite{cai2018cascade} & ResNet101 &  \textbf{58.9} & \textbf{85.9} & \textbf{72.5} & \textbf{52.7} & \textbf{65.7} & \underline{74.5} \\ 
		Grid R-CNN~\cite{lu2019grid} & ResNet101 &  \underline{57.5} & 83.1 & \underline{70.9} & \underline{51.2} & \underline{64.9} & \textbf{75.3} \\ 
		Sparse R-CNN~\cite{sun2021sparse} & ResNet101 & 46.9 & 71.2 & 57.7 & 42.1 & 52.5 & 59.9 \\ 
		\bottomrule
	\end{tabular}
\end{table*}

\section{Experimental Evaluation}
We evaluated some of the more popular object detection models in recent years on WalnutData, and implemented one-stage and two-stage object detection algorithms using the ultralytics framework~\cite{jocher2022ultralytics} and the mmdetection framework~\cite{MMDetection_Contributors_OpenMMLab_Detection_Toolbox_2018} respectively. The one-stage object detection algorithms include YOLOv3~\cite{redmon2018yolov3}, YOLOv4~\cite{bochkovskiy2020yolov4}, YOLOv5~\cite{jocher2022ultralytics}, DETR~\cite{carion2020end}, etc.; the two-stage object detection algorithms include Fast R-CNN~\cite{girshick2015fast}, Faster R-CNN~\cite{ren2015faster}, TridentNet~\cite{li2019scale}, etc. In the following content, the evaluation results of each algorithm on instances of various categories and sizes in WalnutData will be announced. All the experiments of the models in this study were carried out on servers equipped with 8 RTX 3090 GPUs or A800 GPUs, and the hyperparameters of the baseline models all used the default parameter values. In addition, the evaluation results of these baseline models will be provided as benchmark values of WalnutData for researchers as a reference.  

\subsection{Baselines of One-Stage Object Detection Algorithms}
We conducted benchmark evaluations on WalnutData using YOLOv3~\cite{redmon2018yolov3}, YOLOv4~\cite{bochkovskiy2020yolov4}, YOLOv5~\cite{jocher2022ultralytics}, YOLOv6~\cite{li2022yolov6,wang2023yolov7,jocher2022ultralytics,wang2024yolov9,wang2025yolov10} to YOLOv11~\cite{khanam2024yolov11}, as well as YOLOX~\cite{ge2021yolox}, DETR~\cite{carion2020end}, Deformable DETR~\cite{zhu2020deformable}, DINO~\cite{zhang2022dino}, and Conditional DETR~\cite{meng2021conditional}. From the evaluation results on the validation set of WalnutData (Table~\ref{tab:Table_6}), it can be seen that YOLOv3 (154.6) and YOLOv3-SPP (155.4) with relatively high GFLOPs rank first and second in terms of the mAP50 metric. However, in terms of the mAP50:95 metric, YOLOv8x (72.7\%) and YOLOv10x (72.6\%) have better detection performance. Under the mmdetection framework, we used AP50:95, AP50, and AP75 as evaluation metrics. Among the evaluation metric results of these models (Table~\ref{tab:Table_7}), YOLOX shows the strongest performance.

In Table~\ref{tab:Table_8} and Table~\ref{tab:Table_9}, we will present the detection accuracies of these one-stage object detection algorithms for each category and for small, medium, and large-sized targets. In the evaluation results of the benchmark algorithms in Table~\ref{tab:Table_8}, YOLOv3-SPP performs the best in terms of the mAP50 metric. In addition, in the evaluation results of the mAP50:95 metric, YOLOv8x achieves 77.0\%, 73.1\%, 68.8\%, and 71.8\% for the A1, B1, B2, and A2 categories respectively, and its overall strength is the highest among other algorithms. Moreover, as can be seen from Table~\ref{tab:Table_9}, in the detection of small and medium-sized targets, YOLOX-x has the highest AP values, reaching 53.5\% and 56.3\% respectively, followed by YOLOX-l (50.1\%, 53.3\%) and Deformable DETR (43.7\%, 55.4\%). In the detection of large-sized targets, compared with other algorithms, Deformable DETR has a huge advantage in detection performance, and the corresponding AR metric (72.5\%) also ranks second.

\subsection{Baselines of two-stage object detection algorithms}
This study uses several popular two-stage object detection algorithms in recent years: Fast R-CNN~\cite{girshick2015fast}, Faster R-CNN~\cite{ren2015faster}, Cascade R-CNN~\cite{cai2018cascade}, Grid R-CNN~\cite{lu2019grid}, TridentNet~\cite{li2019scale}, Double Head R-CNN~\cite{wu2020rethinking}, and Sparse R-CNN~\cite{sun2021sparse}, and evaluates these algorithms on WalnutData. As shown in Table~\ref{tab:Table_10}, in the evaluation of two-stage object detection algorithms, Cascade R-CNN ranks first in overall performance, with the best detection results for small objects (52.7\%) and medium-sized objects (65.7\%). Grid R-CNN ranks second, with slightly lower detection performance for small and medium-sized objects compared to Cascade R-CNN, achieving 51.2\% and 64.9\%, respectively. However, Grid R-CNN outperforms Cascade R-CNN in detecting large objects.

\section{Conclusion}
This research aims to address the computer vision challenges of walnut fruit detection from a drone perspective, such as the impacts of lighting variations and occlusions on the algorithms. To this end, we have constructed a fine-grained agricultural drone dataset for walnut detection, which is the first large-scale dataset in the field of smart walnut farming. The dataset’s scale and fine-grained feature segmentation give it significant research value and engineering application potential in the field of agricultural computer vision.In addition, by conducting benchmark evaluations of WalnutData using a series of mainstream object detection models, we hope to drive the development of precision agriculture and the smart walnut sector.

In the future, research based on WalnutData can further advance the application of automated harvesting robots and precision management systems in smart agriculture. By optimizing existing object detection algorithms and integrating more agricultural data, more efficient and accurate crop monitoring and yield prediction can be achieved.

\section*{ACKNOWLEDGMENT}
This study is supported by the Yunnan Province Applied Basic Research Program Key Project (202401AS070034) and the Yunnan Province Forest and Grassland Science and Technology Innovation Joint Project (202404CB090002). We thank Haoyu Wang, Shuangyao Liu, Tingfeng Li, Shuyi Wan, Haotian Feng, Luhao Fang, Songfan Shi, Shiyu Du and all the others who involved in the annotations of WalnutData.
\bibliographystyle{plainnat}
\bibliography{BibForWalnutData}

\begin{thebibliography}{44}
\providecommand{\natexlab}[1]{#1}
\providecommand{\url}[1]{\texttt{#1}}
\expandafter\ifx\csname urlstyle\endcsname\relax
  \providecommand{\doi}[1]{doi: #1}\else
  \providecommand{\doi}{doi: \begingroup \urlstyle{rm}\Url}\fi

\bibitem[Apeinans et~al.(2024)Apeinans, Sondors, Litavniece, Kodors, Zarembo,
  and Feldmane]{apeinans2024cherry}
Ilmars Apeinans, Marks Sondors, Lien{\=\i}te Litavniece, Sergejs Kodors, Imants
  Zarembo, and Daina Feldmane.
\newblock Cherry fruitlet detection using yolov5 or yolov8?
\newblock In \emph{ENVIRONMENT. TECHNOLOGIES. RESOURCES. Proceedings of the
  International Scientific and Practical Conference}, volume~2, pages 29--33,
  2024.

\bibitem[Ariza-Sent{\'\i}s et~al.(2024)Ariza-Sent{\'\i}s, V{\'e}lez,
  Mart{\'\i}nez-Pe{\~n}a, Baja, and Valente]{ariza2024object}
Mar Ariza-Sent{\'\i}s, Sergio V{\'e}lez, Raquel Mart{\'\i}nez-Pe{\~n}a, Hilmy
  Baja, and Jo{\~a}o Valente.
\newblock Object detection and tracking in precision farming: A systematic
  review.
\newblock \emph{Computers and Electronics in Agriculture}, 219:\penalty0
  108757, 2024.

\bibitem[Bargoti and Underwood(2017)]{bargoti2017deep}
Suchet Bargoti and James Underwood.
\newblock Deep fruit detection in orchards.
\newblock In \emph{2017 IEEE international conference on robotics and
  automation (ICRA)}, pages 3626--3633. IEEE, 2017.

\bibitem[Bochkovskiy et~al.(2020)Bochkovskiy, Wang, and
  Liao]{bochkovskiy2020yolov4}
Alexey Bochkovskiy, Chien-Yao Wang, and Hong-Yuan~Mark Liao.
\newblock Yolov4: Optimal speed and accuracy of object detection.
\newblock \emph{arXiv preprint arXiv:2004.10934}, 2020.

\bibitem[Butte et~al.(2021)Butte, Vakanski, Duellman, Wang, and
  Mirkouei]{butte2021potato}
Sujata Butte, Aleksandar Vakanski, Kasia Duellman, Haotian Wang, and Amin
  Mirkouei.
\newblock Potato crop stress identification in aerial images using deep
  learning-based object detection.
\newblock \emph{Agronomy Journal}, 113\penalty0 (5):\penalty0 3991--4002, 2021.

\bibitem[Cai and Vasconcelos(2018)]{cai2018cascade}
Zhaowei Cai and Nuno Vasconcelos.
\newblock Cascade r-cnn: Delving into high quality object detection.
\newblock In \emph{Proceedings of the IEEE conference on computer vision and
  pattern recognition}, pages 6154--6162, 2018.

\bibitem[Carion et~al.(2020)Carion, Massa, Synnaeve, Usunier, Kirillov, and
  Zagoruyko]{carion2020end}
Nicolas Carion, Francisco Massa, Gabriel Synnaeve, Nicolas Usunier, Alexander
  Kirillov, and Sergey Zagoruyko.
\newblock End-to-end object detection with transformers.
\newblock In \emph{European conference on computer vision}, pages 213--229.
  Springer, 2020.

\bibitem[Du et~al.(2018)Du, Qi, Yu, Yang, Duan, Li, Zhang, Huang, and
  Tian]{du2018unmanned}
Dawei Du, Yuankai Qi, Hongyang Yu, Yifan Yang, Kaiwen Duan, Guorong Li, Weigang
  Zhang, Qingming Huang, and Qi~Tian.
\newblock The unmanned aerial vehicle benchmark: Object detection and tracking.
\newblock In \emph{Proceedings of the European conference on computer vision
  (ECCV)}, pages 370--386, 2018.

\bibitem[Du et~al.(2023)Du, Cheng, Ma, Lu, Wang, Meng, Jiang, and
  Hong]{du2023dsw}
Xiaoqiang Du, Hongchao Cheng, Zenghong Ma, Wenwu Lu, Mengxiang Wang, Zhichao
  Meng, Chengjie Jiang, and Fangwei Hong.
\newblock Dsw-yolo: A detection method for ground-planted strawberry fruits
  under different occlusion levels.
\newblock \emph{Computers and electronics in agriculture}, 214:\penalty0
  108304, 2023.

\bibitem[Ge et~al.(2021)Ge, Liu, Wang, Li, and Sun]{ge2021yolox}
Zheng Ge, Songtao Liu, Feng Wang, Zeming Li, and Jian Sun.
\newblock Yolox: Exceeding yolo series in 2021.
\newblock \emph{arXiv preprint arXiv:2107.08430}, 2021.

\bibitem[Girshick(2015)]{girshick2015fast}
Ross Girshick.
\newblock Fast r-cnn.
\newblock In \emph{Proceedings of the IEEE international conference on computer
  vision}, pages 1440--1448, 2015.

\bibitem[H{\"a}ni et~al.(2020)H{\"a}ni, Roy, and Isler]{hani2020minneapple}
Nicolai H{\"a}ni, Pravakar Roy, and Volkan Isler.
\newblock Minneapple: a benchmark dataset for apple detection and segmentation.
\newblock \emph{IEEE Robotics and Automation Letters}, 5\penalty0 (2):\penalty0
  852--858, 2020.

\bibitem[Jia et~al.(2024)Jia, Fu, Lan, Wang, and Su]{jia2024maize}
Yinjiang Jia, Kang Fu, Hao Lan, Xiru Wang, and Zhongbin Su.
\newblock Maize tassel detection with ca-yolo for uav images in complex field
  environments.
\newblock \emph{Computers and Electronics in Agriculture}, 217:\penalty0
  108562, 2024.

\bibitem[Jocher et~al.(2022)Jocher, Chaurasia, Stoken, Borovec, Kwon, Michael,
  Fang, Wong, Yifu, Montes, et~al.]{jocher2022ultralytics}
Glenn Jocher, Ayush Chaurasia, Alex Stoken, Jirka Borovec, Yonghye Kwon, Kalen
  Michael, Jiacong Fang, Colin Wong, Zeng Yifu, Diego Montes, et~al.
\newblock ultralytics/yolov5: v6. 2-yolov5 classification models, apple m1,
  reproducibility, clearml and deci. ai integrations.
\newblock \emph{Zenodo}, 2022.

\bibitem[Khanam and Hussain(2024)]{khanam2024yolov11}
Rahima Khanam and Muhammad Hussain.
\newblock Yolov11: An overview of the key architectural enhancements.
\newblock \emph{arXiv preprint arXiv:2410.17725}, 2024.

\bibitem[Kiefer et~al.(2023)Kiefer, Kristan, Per{\v{s}}, {\v{Z}}ust, Poiesi,
  Andrade, Bernardino, Dawkins, Raitoharju, Quan, et~al.]{kiefer20231st}
Benjamin Kiefer, Matej Kristan, Janez Per{\v{s}}, Lojze {\v{Z}}ust, Fabio
  Poiesi, Fabio Andrade, Alexandre Bernardino, Matthew Dawkins, Jenni
  Raitoharju, Yitong Quan, et~al.
\newblock 1st workshop on maritime computer vision (macvi) 2023: Challenge
  results.
\newblock In \emph{Proceedings of the IEEE/CVF Winter Conference on
  Applications of Computer Vision}, pages 265--302, 2023.

\bibitem[Li et~al.(2022)Li, Li, Jiang, Weng, Geng, Li, Ke, Li, Cheng, Nie,
  et~al.]{li2022yolov6}
Chuyi Li, Lulu Li, Hongliang Jiang, Kaiheng Weng, Yifei Geng, Liang Li, Zaidan
  Ke, Qingyuan Li, Meng Cheng, Weiqiang Nie, et~al.
\newblock Yolov6: A single-stage object detection framework for industrial
  applications.
\newblock \emph{arXiv preprint arXiv:2209.02976}, 2022.

\bibitem[Li et~al.(2019)Li, Chen, Wang, and Zhang]{li2019scale}
Yanghao Li, Yuntao Chen, Naiyan Wang, and Zhaoxiang Zhang.
\newblock Scale-aware trident networks for object detection.
\newblock In \emph{Proceedings of the IEEE/CVF international conference on
  computer vision}, pages 6054--6063, 2019.

\bibitem[Liang et~al.(2025)Liang, Liang, Yang, Ge, Zhao, Li, Bai, Fan, Lan, and
  Long]{liang2025enhanced}
Changjiang Liang, Juntao Liang, Weiguang Yang, Weiyi Ge, Jing Zhao, Zhaorong
  Li, Shudai Bai, Jiawen Fan, Yubin Lan, and Yongbing Long.
\newblock Enhanced visual detection of litchi fruit in complex natural
  environments based on unmanned aerial vehicle (uav) remote sensing.
\newblock \emph{Precision Agriculture}, 26\penalty0 (1):\penalty0 23, 2025.

\bibitem[Lin et~al.(2014)Lin, Maire, Belongie, Hays, Perona, Ramanan,
  Doll{\'a}r, and Zitnick]{lin2014microsoft}
Tsung-Yi Lin, Michael Maire, Serge Belongie, James Hays, Pietro Perona, Deva
  Ramanan, Piotr Doll{\'a}r, and C~Lawrence Zitnick.
\newblock Microsoft coco: Common objects in context.
\newblock In \emph{Computer vision--ECCV 2014: 13th European conference,
  zurich, Switzerland, September 6-12, 2014, proceedings, part v 13}, pages
  740--755. Springer, 2014.

\bibitem[Lu et~al.(2019)Lu, Li, Yue, Li, and Yan]{lu2019grid}
Xin Lu, Buyu Li, Yuxin Yue, Quanquan Li, and Junjie Yan.
\newblock Grid r-cnn.
\newblock In \emph{Proceedings of the IEEE/CVF conference on computer vision
  and pattern recognition}, pages 7363--7372, 2019.

\bibitem[Meng et~al.(2021)Meng, Chen, Fan, Zeng, Li, Yuan, Sun, and
  Wang]{meng2021conditional}
Depu Meng, Xiaokang Chen, Zejia Fan, Gang Zeng, Houqiang Li, Yuhui Yuan, Lei
  Sun, and Jingdong Wang.
\newblock Conditional detr for fast training convergence.
\newblock In \emph{Proceedings of the IEEE/CVF international conference on
  computer vision}, pages 3651--3660, 2021.

\bibitem[{MMDetection
  Contributors}(2018)]{MMDetection_Contributors_OpenMMLab_Detection_Toolbox_2018}
{MMDetection Contributors}.
\newblock {OpenMMLab Detection Toolbox and Benchmark}, August 2018.
\newblock URL \url{https://github.com/open-mmlab/mmdetection}.

\bibitem[Quero and Martinez-Carranza(2025)]{quero2025unmanned}
Carlos~Osorio Quero and Jose Martinez-Carranza.
\newblock Unmanned aerial systems in search and rescue: A global perspective on
  current challenges and future applications.
\newblock \emph{International Journal of Disaster Risk Reduction}, page 105199,
  2025.

\bibitem[Redmon and Farhadi(2018)]{redmon2018yolov3}
Joseph Redmon and Ali Farhadi.
\newblock Yolov3: An incremental improvement.
\newblock \emph{arXiv preprint arXiv:1804.02767}, 2018.

\bibitem[Ren et~al.(2015)Ren, He, Girshick, and Sun]{ren2015faster}
Shaoqing Ren, Kaiming He, Ross Girshick, and Jian Sun.
\newblock Faster r-cnn: Towards real-time object detection with region proposal
  networks.
\newblock \emph{Advances in neural information processing systems}, 28, 2015.

\bibitem[Santos and Gebler(2021)]{santos2021methodology}
Thiago~T Santos and Luciano Gebler.
\newblock A methodology for detection and localization of fruits in apples
  orchards from aerial images.
\newblock \emph{arXiv preprint arXiv:2110.12331}, 2021.

\bibitem[Stein et~al.(2016)Stein, Bargoti, and Underwood]{stein2016image}
Madeleine Stein, Suchet Bargoti, and James Underwood.
\newblock Image based mango fruit detection, localisation and yield estimation
  using multiple view geometry.
\newblock \emph{Sensors}, 16\penalty0 (11):\penalty0 1915, 2016.

\bibitem[Sun et~al.(2021)Sun, Zhang, Jiang, Kong, Xu, Zhan, Tomizuka, Li, Yuan,
  Wang, et~al.]{sun2021sparse}
Peize Sun, Rufeng Zhang, Yi~Jiang, Tao Kong, Chenfeng Xu, Wei Zhan, Masayoshi
  Tomizuka, Lei Li, Zehuan Yuan, Changhu Wang, et~al.
\newblock Sparse r-cnn: End-to-end object detection with learnable proposals.
\newblock In \emph{Proceedings of the IEEE/CVF conference on computer vision
  and pattern recognition}, pages 14454--14463, 2021.

\bibitem[Varga et~al.(2022)Varga, Kiefer, Messmer, and
  Zell]{varga2022seadronessee}
Leon~Amadeus Varga, Benjamin Kiefer, Martin Messmer, and Andreas Zell.
\newblock Seadronessee: A maritime benchmark for detecting humans in open
  water.
\newblock In \emph{Proceedings of the IEEE/CVF winter conference on
  applications of computer vision}, pages 2260--2270, 2022.

\bibitem[Wang et~al.(2025{\natexlab{a}})Wang, Chen, Liu, Chen, Lin, Han,
  et~al.]{wang2025yolov10}
Ao~Wang, Hui Chen, Lihao Liu, Kai Chen, Zijia Lin, Jungong Han, et~al.
\newblock Yolov10: Real-time end-to-end object detection.
\newblock \emph{Advances in Neural Information Processing Systems},
  37:\penalty0 107984--108011, 2025{\natexlab{a}}.

\bibitem[Wang et~al.(2023)Wang, Bochkovskiy, and Liao]{wang2023yolov7}
Chien-Yao Wang, Alexey Bochkovskiy, and Hong-Yuan~Mark Liao.
\newblock Yolov7: Trainable bag-of-freebies sets new state-of-the-art for
  real-time object detectors.
\newblock In \emph{Proceedings of the IEEE/CVF conference on computer vision
  and pattern recognition}, pages 7464--7475, 2023.

\bibitem[Wang et~al.(2024)Wang, Yeh, and Mark~Liao]{wang2024yolov9}
Chien-Yao Wang, I-Hau Yeh, and Hong-Yuan Mark~Liao.
\newblock Yolov9: Learning what you want to learn using programmable gradient
  information.
\newblock In \emph{European conference on computer vision}, pages 1--21.
  Springer, 2024.

\bibitem[Wang et~al.(2025{\natexlab{b}})Wang, Yun, Yang, Wu, Wang, and
  Chen]{wang2025ow}
Haoyu Wang, Lijun Yun, Chenggui Yang, Mingjie Wu, Yansong Wang, and Zaiqing
  Chen.
\newblock Ow-yolo: An improved yolov8s lightweight detection method for
  obstructed walnuts.
\newblock \emph{Agriculture}, 15\penalty0 (2):\penalty0 159,
  2025{\natexlab{b}}.

\bibitem[Wen et~al.(2024)Wen, Shi, Wang, Chen, Di, and Yang]{wen2024route}
Haolin Wen, Yuhe Shi, Songyi Wang, Tong Chen, Peng Di, and Lili Yang.
\newblock Route planning for uavs maritime search and rescue considering the
  targets moving situation.
\newblock \emph{Ocean Engineering}, 310:\penalty0 118623, 2024.

\bibitem[Wu et~al.(2024)Wu, Yun, Xue, Chen, and Xia]{wu2024walnut}
Mingjie Wu, Lijun Yun, Chen Xue, Zaiqing Chen, and Yuelong Xia.
\newblock Walnut recognition method for uav remote sensing images.
\newblock \emph{Agriculture}, 14\penalty0 (4):\penalty0 646, 2024.

\bibitem[Wu et~al.(2020)Wu, Chen, Yuan, Liu, Wang, Li, and
  Fu]{wu2020rethinking}
Yue Wu, Yinpeng Chen, Lu~Yuan, Zicheng Liu, Lijuan Wang, Hongzhi Li, and Yun
  Fu.
\newblock Rethinking classification and localization for object detection.
\newblock In \emph{Proceedings of the IEEE/CVF conference on computer vision
  and pattern recognition}, pages 10186--10195, 2020.

\bibitem[Wu et~al.(2023)Wu, Liu, Sun, and Wang]{wu2023dataset}
Zhen-wei Wu, Ming-hao Liu, Cheng-xiu Sun, and Xin-fa Wang.
\newblock A dataset of tomato fruits images for object detection in the complex
  lighting environment of plant factories.
\newblock \emph{Data in Brief}, 48:\penalty0 109291, 2023.

\bibitem[Yang et~al.(2025)Yang, Chen, Chen, Ma, Chen, and Zhao]{yang2025walnut}
Rui Yang, Dan Chen, Yanling Chen, Yage Ma, Chaoyin Chen, and Shenglan Zhao.
\newblock Walnut oil prevents hyperlipidemia induced by high-fat diet and
  regulates intestinal flora and liver metabolism.
\newblock \emph{Frontiers in Pharmacology}, 15:\penalty0 1431649, 2025.

\bibitem[Zhang et~al.(2022)Zhang, Li, Liu, Zhang, Su, Zhu, Ni, and
  Shum]{zhang2022dino}
Hao Zhang, Feng Li, Shilong Liu, Lei Zhang, Hang Su, Jun Zhu, Lionel~M Ni, and
  Heung-Yeung Shum.
\newblock Dino: Detr with improved denoising anchor boxes for end-to-end object
  detection.
\newblock \emph{arXiv preprint arXiv:2203.03605}, 2022.

\bibitem[Zheng et~al.(2024)Zheng, Wu, Chen, Wang, Xiong, Wei, Huang, Wang,
  Huang, and Du]{zheng2024robust}
Zhenhui Zheng, Meng Wu, Ling Chen, Chenglin Wang, Juntao Xiong, Lijiao Wei,
  Xiaoman Huang, Shuo Wang, Weihua Huang, and Dongjie Du.
\newblock A robust and efficient citrus counting approach for large-scale
  unstructured orchards.
\newblock \emph{Agricultural Systems}, 215:\penalty0 103867, 2024.

\bibitem[Zhu et~al.(2021)Zhu, Wen, Du, Bian, Fan, Hu, and
  Ling]{zhu2021detection}
Pengfei Zhu, Longyin Wen, Dawei Du, Xiao Bian, Heng Fan, Qinghua Hu, and Haibin
  Ling.
\newblock Detection and tracking meet drones challenge.
\newblock \emph{IEEE transactions on pattern analysis and machine
  intelligence}, 44\penalty0 (11):\penalty0 7380--7399, 2021.

\bibitem[Zhu et~al.(2025)Zhu, Niu, Yue, and Zhou]{zhu2025multiscale}
Wenyu Zhu, Shanwei Niu, Jixiang Yue, and Yangli Zhou.
\newblock Multiscale wildfire and smoke detection in complex drone forest
  environments based on yolov8.
\newblock \emph{Scientific Reports}, 15\penalty0 (1):\penalty0 2399, 2025.

\bibitem[Zhu et~al.(2020)Zhu, Su, Lu, Li, Wang, and Dai]{zhu2020deformable}
Xizhou Zhu, Weijie Su, Lewei Lu, Bin Li, Xiaogang Wang, and Jifeng Dai.
\newblock Deformable detr: Deformable transformers for end-to-end object
  detection.
\newblock \emph{arXiv preprint arXiv:2010.04159}, 2020.

\end{thebibliography}

\end{document}